\documentclass{article} % For LaTeX2e
\usepackage{iclr2027_conference,times}

\usepackage{amsmath,amsfonts,bm}

\def\eqref#1{equation~\ref{#1}}
\def\1{\bm{1}}

\DeclareMathAlphabet{\mathsfit}{\encodingdefault}{\sfdefault}{m}{sl}
\SetMathAlphabet{\mathsfit}{bold}{\encodingdefault}{\sfdefault}{bx}{n}

\usepackage{hyperref}
\usepackage{url}
\usepackage{booktabs}
\usepackage{graphicx}
\usepackage{subcaption}
\usepackage{algorithm}
\usepackage{float}
\usepackage{algpseudocode}
\usepackage{tcolorbox}
\tcbuselibrary{listings,skins,breakable}
\newtcblisting{promptbox}[1]{
    enhanced,
    breakable,
    title={#1},
    colback=white,
    colbacktitle=gray!15,
    coltitle=black,
    colframe=black!70,
    fonttitle=\bfseries,
    boxrule=0.5pt,
    arc=2mm,
    left=3mm, right=3mm, top=2mm, bottom=2mm,
    listing only,
    listing options={
        basicstyle=\ttfamily\small,
        columns=fullflexible,
        keepspaces=true,
        breaklines=true,
        breakatwhitespace=true,
        showstringspaces=false
    }
}
\title{SAPD: Step-Aligned Privileged Distillation}

\author{\small
\hspace{-0.3em}Tianle Wang\textsuperscript{1,2*}\hspace{0.3em}
Jiayu Liu\textsuperscript{2*}\hspace{0.3em}
Ruizhi Zhao\textsuperscript{2,3}\hspace{0.3em}
Ning Miao\textsuperscript{1,2}
}

\newcommand{\model}{SAPD}

\iclrfinalcopy % Uncomment for camera-ready version, but NOT for submission.

\begin{document}

\maketitle
\footnotetext[1]{Department of Data Science, City University of Hong Kong,~
\textsuperscript{2}Hong Kong Institute of AI for Science, City University of Hong Kong,~
\textsuperscript{3}Shenzhen University,~
\textsuperscript{*}Equal contribution. Correspondence to: Ning Miao (ningmiao@cityu.edu.hk).}

\lhead{Preprint}

\begin{abstract}
On-policy post-training can improve large language models by learning from their own trajectories, but requires costly rollout generation.
We ask whether fixed demonstrations can support competitive off-policy learning through better supervision.
Our premise is that their usefulness depends not only on the training trajectories, but also on whether supervision provides informative preferences among continuations and connects this guidance to the reasoning decision being learned.
We introduce \textbf{Step-Aligned Privileged Distillation (\model{})}, a rollout-free self-distillation method that turns demonstrations into step-aligned distributional supervision.
Its key insight is to use the known progression of a reference solution to associate each reasoning transition with targeted privileged guidance, rather than treating the solution as undifferentiated context.
On mathematical reasoning benchmarks, \model{} outperforms supervised fine-tuning and label smoothing on average while remaining competitive with on-policy reinforcement learning and self-distillation.
Analyses support both the value of context-dependent distributional guidance and the benefit of aligning privileged information with the current step.
\model{} also largely preserves out-of-domain coding performance and achieves approximately $2\times$ training-loop speedups over the on-policy baselines.
These findings suggest that carefully constructed supervision can make fully off-policy post-training a competitive and computationally efficient alternative. 
Our code is available at \url{https://github.com/Miaow-Lab/SAPD}.
\end{abstract}

\section{Introduction}
Post-training enables large language models (LLMs) to acquire specialized capabilities from task-specific data and feedback. A central choice is whether to learn from fixed demonstrations or from trajectories generated by the model being trained.
On-policy approaches include reinforcement learning with verifiable rewards, such as Group Relative Policy Optimization (GRPO)~\citep{shao2024deepseekmathpushinglimitsmathematical}, and distillation from teacher feedback on student-generated trajectories~\citep{agarwal2024onpolicy}. Recent self-distillation methods further use demonstrations to guide on-policy learning without requiring a separately trained, larger teacher~\citep{zhao2026selfdistilled,shenfeld2026selfdistillation}.
A common motivation is that learning on the student's own trajectories reduces the mismatch between training and inference, which is believed to contribute to strong task performance and generalization~\citep{agarwal2024onpolicy}. However, repeatedly generating these trajectories introduces substantial autoregressive decoding costs, particularly for long reasoning traces~\citep{shenfeld2026selfdistillation}.

This trade-off raises a practical question: \textbf{can fixed demonstrations support competitive post-training without student rollouts?}
% 对比on-off policy，有2个主要的区别
% Our approach is motivated by two connected principles: \emph{informative distributional supervision} and \emph{step-aligned privileged information}.
We study this question from two complementary perspectives.
First, rather than treating each demonstrated token as the complete supervision signal, we ask whether fixed trajectories can provide richer, distributional guidance over possible next-token predictions.
Second, because a reasoning demonstration contains a sequence of intermediate steps, we ask whether such guidance should be localized to the step associated with the current prediction rather than derived from the entire solution.
Together, these perspectives examine whether the effectiveness of fixed demonstrations can be improved by changing how supervision is constructed, without changing the trajectories on which learning takes place.

First, \textbf{supervision should express informative preferences, not merely a demonstrated token or uniformly softened labels}.
A reasoning trace combines decisions needed to solve a problem with choices of wording and exposition. One-hot imitation does not explicitly distinguish plausible alternative expressions from incompatible continuations.
Knowledge distillation provides a high-level alternative: a teacher's predictive distribution can convey relative support beyond a single observed label~\citep{hinton2015distillingknowledgeneuralnetwork}. Privileged-information learning further suggests that information available only during training can be translated into such supervision~\citep{lopezpaz2016unifyingdistillationprivilegedinformation}.
We therefore hypothesize that demonstration-informed distributions can guide learning on fixed trajectories more effectively than hard targets or uniform smoothing, by expressing context-dependent preferences and uncertainty.

Second, \textbf{privileged guidance should be aligned with the reasoning decision being learned}.
Distributional targets provide a supervision mechanism, but their usefulness also depends on which part of the demonstration is used to construct them.
Work on process supervision motivates connecting feedback to intermediate reasoning decisions rather than relying solely on final outcomes~\citep{uesato2022solvingmathwordproblems,ICLR2024_aca97732}.
We apply this intuition to the scope of privileged information, rather than to step-level reward labels. Giving a teacher the full solution leaves it to determine which parts should guide each local continuation. Fixed demonstrations offer a useful alternative: their known progression identifies the reasoning step associated with each training context.
We hypothesize that making this correspondence explicit provides more focused guidance than using the same global solution context throughout the trace.

These principles lead to Step-Aligned Privileged Distillation (\model{}), a rollout-free self-distillation method.
\model{} uses a demonstration's step structure to associate each reference transition with corresponding privileged guidance, then distills the resulting context-dependent distributions on fixed reference trajectories.
Distributional supervision determines \emph{what preferences are transferred}; step alignment determines \emph{which demonstrated information informs those preferences}. Their combination turns fixed demonstrations into targeted learning signals without generating new training trajectories.

\model{} achieves strong average performance across three mathematical reasoning benchmarks, outperforming SFT and label-smoothed SFT while remaining competitive with GRPO and on-policy self-distillation (OPSD)~\citep{zhao2026selfdistilled}. It also largely preserves out-of-domain coding performance.
We then examine the two motivating hypotheses: target-distribution examples and training dynamics assess informative supervision beyond generic softness, while full-solution and misaligned-guidance ablations assess the value of step alignment.
Finally, our efficiency evaluation reports approximately $2\times$ training-loop speedups over the on-policy baselines.
Together, these results suggest that the construction of supervision is an important lever for effective off-policy post-training, alongside the choice of training trajectories.

\section{Preliminaries}
\label{sec:preliminaries}

Let $\mathcal{D}=\{(x,y^\star)\}$ denote a dataset of problems and reference solutions, and let $\pi_\theta$ be an autoregressive student language model over vocabulary $\mathcal{V}$. We use $\theta$ and $\phi$ for student and teacher parameters, respectively, and $\mathrm S$ and $\mathrm T$ to label their predictive distributions. The index $t$ denotes a token position within a sequence. For a context $c$ and temperature $T>0$, define
\begin{equation}
\pi_\psi^{(T)}(\cdot\mid c)
=\operatorname{softmax}\!\left(z_\psi(c)/T\right),
\qquad \psi\in\{\theta,\phi\},
\end{equation}
where $z_\psi(c)$ denotes the next-token logits. We omit the superscript when $T=1$, so $\pi_\psi=\pi_\psi^{(1)}$.

\paragraph{SFT and label smoothing.}
Supervised fine-tuning (SFT) is an off-policy method that learns from fixed reference trajectories through teacher forcing:
\begin{equation}
\mathcal{L}_{\mathrm{SFT}}(\theta)
=
-\mathbb{E}_{(x,y^\star)\sim\mathcal{D}}
\left[
\frac{1}{|y^\star|}
\sum_{t=1}^{|y^\star|}
\log \pi_\theta(y_t^\star \mid x,y_{<t}^\star)
\right].
\end{equation}
Label smoothing~\citep{3295222.3295349} replaces the one-hot target with
$q_t(v)=(1-\epsilon_{\mathrm{ls}})\mathbf{1}[v=y_t^\star]
+\frac{\epsilon_{\mathrm{ls}}}{|\mathcal{V}|-1}\mathbf{1}[v\neq y_t^\star]$
and minimizes cross-entropy against $q_t$.
Both methods train on reference prefixes; label smoothing softens the supervision but assigns equal probability to all alternative tokens.

\paragraph{GRPO.}
Group Relative Policy Optimization (GRPO)~\citep{shao2024deepseekmathpushinglimitsmathematical} is an on-policy method commonly used for reinforcement learning with verifiable rewards.
For each problem, it samples $G$ responses from the rollout policy $\pi_{\theta_{\mathrm{old}}}$ and obtains verifiable rewards $\{r_i\}_{i=1}^{G}$.
Each response receives a group-normalized advantage
$\hat{A}_i=(r_i-\bar{r})/(\sigma_r+\varepsilon)$,
where $\bar{r}$ and $\sigma_r$ are the group reward mean and standard deviation.
GRPO optimizes a clipped policy-gradient objective, optionally with a reference-policy KL penalty, without a learned value model.
Under outcome supervision, all tokens within a response share the same advantage.

\paragraph{OPSD.}
On-policy self-distillation (OPSD)~\citep{zhao2026selfdistilled} provides dense distributional supervision along student-generated trajectories $\hat{y}\sim\pi_\theta(\cdot\mid x)$.
A teacher derived from the same base model additionally observes the reference solution as privileged information.
At each student prefix, define
$p_{\mathrm S,t}^{(T)}=\pi_\theta^{(T)}(\cdot\mid x,\hat{y}_{<t})$
and
$p_{\mathrm T,t}^{(T)}=\pi_\phi^{(T)}(\cdot\mid x,y^\star,\hat{y}_{<t})$.
Here $T>0$ is the shared distillation temperature. Omitting divergence clipping, its forward-KL objective is
\begin{equation}
\mathcal{L}_{\mathrm{OPSD}}(\theta)
=
\mathbb{E}_{\substack{(x,y^\star)\sim\mathcal{D}\\
\hat{y}\sim\pi_\theta(\cdot\mid x)}}
\left[
\frac{1}{|\hat{y}|}
\sum_{t=1}^{|\hat{y}|}
D_{\mathrm{KL}}\!\left(p_{\mathrm T,t}^{(T)} \,\|\, p_{\mathrm S,t}^{(T)}\right)
\right].
\end{equation}
Gradients flow only through the student predictions.
SDFT~\citep{shenfeld2026selfdistillation} shares the same core paradigm of on-policy distillation from a demonstration-conditioned teacher.
\section{Step-Aligned Privileged Distillation}
\label{sec:method}
We introduce \model{}, a rollout-free self-distillation method organized around two design choices: \emph{the target should convey richer information} and \emph{the information used to construct that target should be relevant to the reasoning decision being learned}. 
This leads to a simple construction: we use a privileged teacher to convert demonstrated reasoning into informative target distributions, while using the progression of the reference solution to determine which privileged information is relevant at each prediction step. Importantly, our method changes the supervision available at fixed reference contexts rather than replacing them with student-generated states.

\paragraph{Reference-anchored learning states.}
We retain the reference-prefix contexts used by SFT, avoiding the need to sample student trajectories. Using the notation of Section~\ref{sec:preliminaries}, we partition each reference solution as $y^\star=g_1\oplus\cdots\oplus g_K$, where $\oplus$ denotes concatenation. Each step $g_i$ is a contiguous reference segment, and the partition preserves the original token sequence. The formulation permits different segmentation granularities, including the unsegmented case $K=1$. For the $t$-th token of step $g_i$, the student's prediction context is
\begin{equation}
    h_{i,t}
    =
    \bigl(x,g_{<i},g_{i,<t}\bigr),
    \label{eq:offline_context}
\end{equation}
where $g_{<i}=g_1\oplus\cdots\oplus g_{i-1}$ contains all preceding reference steps, and $g_{i,<t}$ is the reference prefix within the current step. Here $i$ indexes reference steps and $t$ indexes tokens within step $i$; the pair $(i,t)$ corresponds to the full-sequence token position $\sum_{j<i}|g_j|+t$ used in Section~\ref{sec:preliminaries}.

\paragraph{Step-conditioned distributional targets.}
The remaining question is how to construct an informative target distribution for each reference context.
Conditioning the teacher on the entire solution provides abundant privileged information, but does not explicitly associate that information with the particular reasoning transition currently being learned.
Motivated by the local correspondence between supervision and intermediate reasoning decisions in process supervision~\citep{uesato2022solvingmathwordproblems,ICLR2024_aca97732}, we exploit the known structure of the reference trajectory: for each token position in step $g_i$, the distillation target is obtained from a teacher conditioned on the corresponding reference step $g_i$ as privileged information.
Specifically, the student predicts from $h_{i,t}$, while the teacher additionally conditions on $g_i$. Let $c^{\mathrm T}_{i,t}=(h_{i,t},g_i)$ denote this privileged teacher context. At a shared temperature $T>0$, their distributions are
\begin{equation}
p_{\mathrm S,i,t}^{(T)}
=\pi_\theta^{(T)}(\cdot\mid h_{i,t}),
\qquad
p_{\mathrm T,i,t}^{(T)}
=\pi_\phi^{(T)}(\cdot\mid c^{\mathrm T}_{i,t}).
\end{equation}
The teacher distribution therefore converts the demonstrated step into token-level preferences at the student context.
Crucially, the step boundary controls only the teacher-side privileged information: the student's causal history remains the complete reference prefix and is never reset between steps.

\paragraph{Distilling the aligned distributions.}
We train the student to match the step-conditioned teacher distribution at each reference-token position.
Let $\mathcal I(x,y^\star)$ denote the set of reference-token positions in the demonstration, and let $N=|\mathcal I(x,y^\star)|$. The token-mean forward KL loss is
\begin{equation}
\label{eq:loss}
\mathcal{L}_{\mathrm{\model{}}}(\theta;x,y^\star)
=
\frac{1}{N}\sum_{(i,t)\in\mathcal I(x,y^\star)}
D_{\mathrm{KL}}
\left(
p_{\mathrm T,i,t}^{(T)}
\,\Vert\,
p_{\mathrm S,i,t}^{(T)}
\right).
\end{equation}
The teacher is stop-gradient and remains frozen at the initial checkpoint throughout training.

Algorithm~\ref{alg:sapd} summarizes training. A single causal student forward pass scores the retained reference sequence for each example; teacher forwards are performed separately for each segment. Both branches use teacher forcing, so no autoregressive rollout generation is required.
\begin{algorithm}[t]
\caption{Step-Aligned Privileged Distillation}
\label{alg:sapd}
\begin{algorithmic}[1]
\Require Demonstrations $\mathcal D$, initial parameters $\theta_0$, temperature $T>0$
\State Initialize student $\theta\gets\theta_0$ and frozen teacher $\phi\gets\theta_0$
\While{training has not converged}
    \State Sample a minibatch $\mathcal B\subset\mathcal D$
    \For{each $(x,y^\star)\in\mathcal B$}
        \State Partition $y^\star$ into reference segments $(g_1,\ldots,g_K)$
        \For{each reference position $(i,t)\in\mathcal I(x,y^\star)$}
            \State Form student context $h_{i,t}=(x,g_{<i},g_{i,<t})$
            \State Construct teacher context $c^{\mathrm T}_{i,t}=(h_{i,t},g_i)$
            \State Compute $p_{\mathrm S,i,t}^{(T)}=\operatorname{softmax}(z_\theta(h_{i,t})/T)$
            \State Compute $p_{\mathrm T,i,t}^{(T)}=\operatorname{softmax}(z_\phi(c^{\mathrm T}_{i,t})/T)$
        \EndFor
    \EndFor
    \State Compute the token-mean forward KL loss in Eq.~\ref{eq:loss} over $\mathcal B$
    \State Update $\theta$ by gradient descent while keeping $\phi$ fixed
\EndWhile
\State \Return Student $\pi_\theta$, using no privileged information at inference
\end{algorithmic}
\end{algorithm}

Overall, \model{} uses the reference trajectory for two complementary purposes: its prefixes define rollout-free training states, while its step structure identifies the privileged information used to construct distributional supervision at those states.

\section{Experiments}
\label{sec:experiments}
We first ask whether \model{} achieves competitive reasoning performance while retaining out-of-domain capabilities.
We then examine the two hypotheses motivating the method: whether its supervision provides informative distributional structure beyond generic softness, and whether privileged information is more effective when aligned with the current reasoning step.
Finally, we evaluate the computational consequence of learning without student rollouts.

\subsection{Experiment setup}
\paragraph{Models and datasets.}
We use \texttt{Qwen3-4B-Instruct-2507} as the base model and conduct post-training on mathematical reasoning subset of OpenThoughts~\citep{guha2026openthoughts}. We evaluate in-domain mathematical reasoning on AIME24/25~\citep{aime24, aime25}, and HMMT25~\citep{dekoninck2026matharena}. To assess out-of-domain capability retention after post-training, we additionally evaluate coding performance on MBPP+~\citep{austin2021programsynthesislargelanguage}, HumanEval+~\citep{chen2021evaluatinglargelanguagemodels}, and LiveCodeBench v6 (LCBv6)~\citep{ICLR2025_94074dd5}.

\paragraph{Baselines.}
We compare \model{} against four baselines covering off-policy supervised learning, on-policy reinforcement learning, and on-policy self-distillation. Their formulations are described in Section~\ref{sec:preliminaries}.
(1) \textbf{SFT} serves as the standard supervised baseline.
(2) \textbf{LS-SFT} uses a smoothing coefficient of $\epsilon_{\mathrm{ls}}=0.1$ to test whether the gains of \model{} can be explained by target softening alone.
(3) \textbf{GRPO} serves as a representative reinforcement learning with verifiable rewards (RLVR) baseline, enabling comparison with on-policy learning from outcome-level feedback.
(4) \textbf{OPSD} enables comparison with on-policy distillation using reference solutions as privileged information. Since SDFT follows the same core training paradigm, we use OPSD as the representative baseline for this family and do not evaluate SDFT separately.

\paragraph{Implementation details.}
We remove examples with empty problems, reasoning traces, or solutions. For each remaining example, we concatenate \texttt{COT\_Reason} and \texttt{solution} with a blank-line separator and partition the resulting reference into ordered segments. Segmentation prioritizes explicit step headings, followed by numbered-step markers, and falls back to paragraph boundaries.
Both the student and teacher use the non-thinking chat template with the assistant prefill ``Let's solve the problem step by step.''
To reduce redundant computation, we tokenize the reference as a continuous sequence and share a single causal student forward pass across all segments within each example.
Training and evaluation are performed on four NVIDIA H100 GPUs. OPSD and \model{} use a distillation temperature of $T=1.1$ for both student and teacher logits.

\paragraph{Evaluation settings.}
We follow the evaluation protocol of \citet{zhao2026selfdistilled} and report avg@$k$, the mean fraction of correct responses across $k$ samples per problem, averaged over the benchmark. Here, we set $k=12$.
For each method and base model, we select the checkpoint with the highest mean avg@$k$ across the in-domain benchmarks.
Sampling settings, prompting, and answer grading are detailed in Appendix~\ref{app:evaluation_details}.

\subsection{Overall performance and capability generalization}
\begin{table}[t]
\centering
\caption{\textbf{Overall Avg@12 and average output length}. Mathematics benchmarks measure in-domain performance; coding benchmarks measure out-of-domain (OOD) capability retention after mathematics post-training. Average length is measured in tokens on mathematics benchmarks. Among post-trained methods, \textbf{bold} and \underline{underline} indicate the best and second-best results, respectively. Base is shown separately as the initial-model reference.}
\label{tab:overall_performance}
\setlength{\tabcolsep}{4pt}
\renewcommand{\arraystretch}{1.12}
\resizebox{\linewidth}{!}{%
\begin{tabular}{lccccccccc}
\toprule
& \multicolumn{5}{c}{In-domain: Mathematics} & \multicolumn{4}{c}{Out-of-domain: Code} \\
\cmidrule(lr){2-6}\cmidrule(lr){7-10}
Method & AIME24 & AIME25 & HMMT25 & Avg. & Avg. length & MBPP+ & HumanEval+ & LCBv6 & Avg. \\
\midrule
Base & 62.22 & 46.94 & 29.17 & 46.11 & 10,595 & 64.15 & 81.86 & 24.24 & 56.75 \\
\midrule
SFT & 58.33 & 46.67 & 28.61 & 44.54 & 12,835 & 63.23 & 80.34 & 18.32 & 53.96 \\
LS-SFT & 59.72 & \textbf{49.17} & \underline{31.11} & 46.67 & \underline{10,102} & 61.97 & 81.71 & 22.14 & 55.27 \\
GRPO & \underline{62.22} & 47.50 & 30.56 & 46.76 & 10,145 & \textbf{64.95} & \underline{82.62} & \textbf{24.05} & \textbf{57.21} \\
OPSD & 61.94 & 47.78 & \textbf{32.78} & \underline{47.50} & 10,870 & 63.82 & \textbf{83.23} & \underline{23.66} & \underline{56.90} \\
\midrule
\textbf{\model{} (Ours)} & \textbf{65.28} & \underline{48.06} & 30.28 & \textbf{47.87} & \textbf{9,927} & \underline{64.15} & \underline{82.62} & 23.28 & 56.68 \\
\bottomrule
\end{tabular}%
}
\end{table}

Table~\ref{tab:overall_performance} compares \model{} with the baselines across the in-domain and out-of-domain benchmarks.

\paragraph{In-domain mathematical reasoning.}
\model{} achieves the strongest overall mathematical reasoning performance among the evaluated methods, with an average accuracy of 47.87\%. The improvement is consistent with the main motivation of the method: enriching supervision on fixed reference trajectories can recover the benefits typically associated with on-policy training without requiring student-generated trajectories.
Notably, this gain is not accompanied by longer generations.
\model{} produces the shortest average outputs among the post-trained methods (9,927 tokens), while still outperforming both the supervised baselines and the evaluated on-policy methods on average.
This suggests that the improvement is not simply attributable to increased test-time reasoning length, but to more effective post-training supervision.

\paragraph{Out-of-domain generalization.}
The gains in mathematical reasoning do not come at the cost of a substantial degradation in the evaluated coding capabilities.
After mathematics post-training, \model{} retains an OOD average of 56.68\%, outperforming the off-policy SFT and LS-SFT baselines and remaining close to the on-policy GRPO and OPSD baselines.
Together, the in-domain and OOD results show that \model{} improves mathematical reasoning while largely preserving capabilities outside the training domain.

\subsection{Beyond label smoothing: context-dependent supervision}
Our first hypothesis concerns the \emph{information expressed by the targets}.
SFT and LS-SFT provide useful comparisons: one concentrates supervision on the demonstrated token, while the other assigns uniform mass to alternatives.
We examine whether \model{} instead supplies context-dependent preferences and whether student training preserves this structure relative to a shared teacher.

\paragraph{Target distributions distinguish flexibility from precision.}

\begin{figure}[t]
    \centering
    \includegraphics[width=\textwidth]{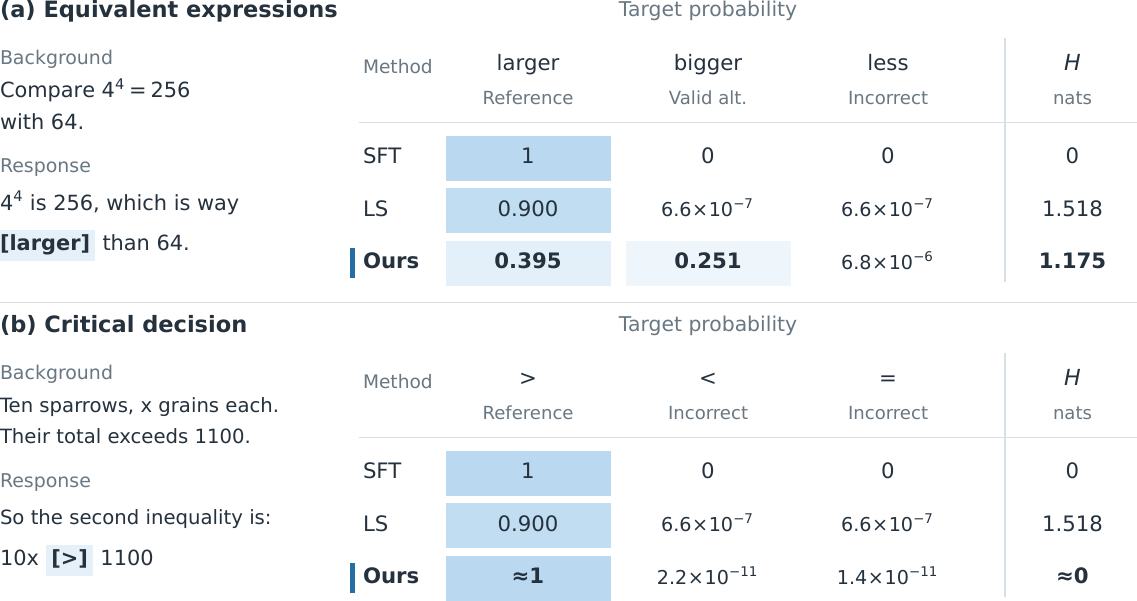}
    \caption{
        \textbf{Informative supervision is not simply target softening.}
            Comparison of SFT, label-smoothed SFT (LS), and \model{} at
            two reference-token positions: (a) equivalent expressions and
            (b) a critical inequality decision. Background gives brief problem context; Response shows a local response, with highlighted tokens marking prediction sites. Cell shading uses a shared linear probability scale from 0 to 1.
            $H$ denotes full-vocabulary entropy.}
    \label{fig:case_study}
\end{figure}

Figure~\ref{fig:case_study} compares the supervision at two reference-token positions.
In the equivalent-expression case, the reference uses \texttt{larger} to compare $4^4=256$ with $64$.
The alternative \texttt{bigger} preserves the comparison, whereas \texttt{less} contradicts it.
SFT assigns no target mass to either alternative, and LS-SFT assigns both the same negligible probability, $6.6\times10^{-7}$.
In contrast, \model{} assigns $0.395$ to \texttt{larger} and $0.251$ to \texttt{bigger}, but only $6.8\times10^{-6}$ to \texttt{less}.
Its supervision therefore distinguishes the alternatives rather than merely spreading probability away from the reference token.

The inequality case validates the complementary requirement of precision.
When ten sparrows each peck $x$ grains and their total exceeds $1100$, the correct relation is $10x>1100$.
Here, \model{} assigns approximately unit probability to $>$, versus $2.2\times10^{-11}$ and $1.4\times10^{-11}$ to $<$ and $=$.
Its target entropy is approximately zero, compared with $1.175$ in the equivalent-expression case.
SFT has zero target entropy in both examples, while LS-SFT has the same reference probability ($0.900$) and entropy ($1.518$) in both.
These examples illustrate the intended property: uncertainty adapts to the prediction context instead of being uniformly increased.

\paragraph{Training dynamics separate aggregate matching from preferences over alternatives.}
For all methods and checkpoints, we evaluate identical fixed reference-prefix contexts against a shared, frozen privileged anchor teacher at temperature $T=1$.

\begin{figure}[t]
    \centering
    \includegraphics[width=\textwidth]{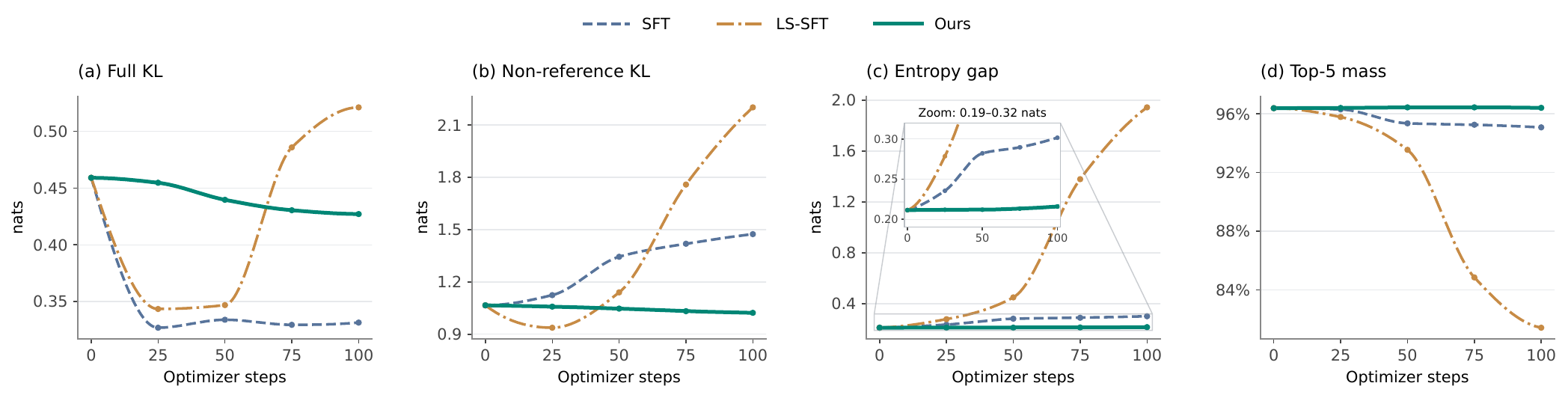}
    \caption{
        \textbf{Training dynamics reveal preservation of non-reference
        distributional structure.}
        (a) Full teacher--student KL,
        $D_{\mathrm{KL}}(p_T \Vert p_S)$.
        (b) Conditional non-reference KL, computed after excluding
        the demonstrated token and renormalizing both distributions.
        (c) Mean token-level absolute entropy gap,
        $|H(p_S)-H(p_T)|$.
        The inset magnifies the range 0.19--0.32.
        (d) Student probability mass on the teacher's top-five tokens.}
    \vspace{-10pt}
    \label{fig:training_dynamics}
\end{figure}

Figure~\ref{fig:training_dynamics}(a) shows that SFT rapidly reduces the full teacher--student KL, $D_{\mathrm{KL}}(p_T \Vert p_S)$, from approximately $0.46$ to $0.33$, followed by a plateau, whereas \model{} produces a smaller reduction to approximately $0.43$.
Label smoothing initially reduces this divergence but subsequently reverses direction, reaching approximately $0.52$.
Thus, SFT achieves the strongest aggregate teacher matching under this metric.
However, aggregate KL combines agreement on the probability assigned to the demonstrated token with agreement on the distribution over all remaining tokens.
To validate the specific hypothesis about preferences beyond the reference token, we separately examine the remaining vocabulary.
For a reference token $y$, define
\begin{equation}
    \bar p_T(v)=\frac{p_T(v)}{1-p_T(y)},
    \qquad
    \bar p_S(v)=\frac{p_S(v)}{1-p_S(y)},
    \qquad v\ne y.
\end{equation}
Conditional KL, $D_{\mathrm{KL}}(\bar p_T\Vert\bar p_S)$, measures relative preferences among non-reference tokens independently of their total probability mass.
In Figure~\ref{fig:training_dynamics}(b), it increases from approximately $1.07$ to $1.47$ under SFT and to $2.20$ under LS-SFT, but decreases modestly to $1.02$ under \model{}.
Writing $a_T=p_T(y)$ and $a_S=p_S(y)$, the KL chain rule explains why the aggregate and conditional trends can differ:
\begin{equation}
\begin{aligned}
D_{\mathrm{KL}}(p_T\Vert p_S)
&=
a_T\log\frac{a_T}{a_S}
+(1-a_T)\log\frac{1-a_T}{1-a_S}\\
&\quad +(1-a_T)
D_{\mathrm{KL}}(\bar p_T\Vert\bar p_S).
\end{aligned}
\label{eq:kl_reference_decomposition}
\end{equation}
Full KL downweights conditional disagreement when the teacher assigns little mass to alternatives.
Accordingly, SFT's lower aggregate KL does not imply better preservation of their relative probabilities.

\paragraph{Uncertainty and candidate-set concentration provide complementary evidence.}
The mean absolute entropy gap, $|H(p_S)-H(p_T)|$, remains near $0.21$ under \model{}, compared with approximately $0.30$ under SFT and $1.95$ under LS-SFT (Figure~\ref{fig:training_dynamics}(c)).
Student mass on the anchor teacher's top-five tokens remains approximately $96.3\%$ under \model{}, but decreases to $95.0\%$ under SFT and $81.4\%$ under LS-SFT (Figure~\ref{fig:training_dynamics}(d)).
These metrics respectively assess uncertainty mismatch and concentration on teacher-preferred candidates.

Together with the target examples, these dynamics support the first hypothesis in a specific sense: \model{} supplies context-dependent preferences and largely preserves non-reference distributional structure that hard imitation and uniform smoothing do not preserve in this comparison.

\subsection{Value of step-aligned privileged information}
\begin{table}[t]
    \centering
    \caption{
        Ablation of privileged supervision.
        We report Avg@12 on three math benchmarks and their average. Best results are in bold.
    }
    \label{tab:ablation}
    \small
    \setlength{\tabcolsep}{8pt}
    \begin{tabular}{lcccc}
        \toprule
        Method & AIME24 & AIME25 & HMMT25 & Avg. \\
        \midrule
        \model{} (ours)
            & \textbf{65.28}
            & 48.06
            & \textbf{30.28}
            & \textbf{47.87} \\

        \quad \textit{Misaligned PI}
            & 60.00
            & \textbf{48.89}
            & 28.06
            & 45.65 \\

        \quad \textit{Full PI}
            & 60.00
            & 44.72
            & 30.27
            & 45.00 \\
        \bottomrule
    \end{tabular}
\end{table}
The second hypothesis concerns \emph{which information constructs the teacher's targets}.
To examine it, we retain the off-policy student reference contexts and distillation objective while comparing current-step guidance with full-solution and misaligned privileged information (PI).

\paragraph{Does broader privileged context improve supervision?}
The \textit{Full PI} variant treats the complete reference solution as one step, so the teacher receives the full solution throughout the trace.
Average accuracy decreases from 47.87\% to 45.00\%, a reduction of 2.87 pp (Table~\ref{tab:ablation}).
Thus, providing more extensive privileged context does not recover the performance of current-step guidance in this comparison.
This supports organizing privileged information around the transition being learned rather than assuming that the complete solution is always the most effective teacher context.

\paragraph{Does the guidance need to correspond to the current step?}
The \textit{Misaligned PI} variant shuffles reference segments within the same problem. For a reference with $K>1$ segments, we use a fixed, seeded permutation $\sigma$ with no fixed points ($\sigma(i)\ne i$), and provide $g_{\sigma(i)}$ as privileged guidance when supervising $g_i$. The student's reference sequence, causal prefixes, and supervised positions remain unchanged; only the assignment of teacher-side guidance is permuted.
Its average accuracy is 45.65\%, or 2.22 pp below \model{}.
Combined with the full-solution comparison, this result supports the value of matching privileged information to the current reference transition.

\subsection{Efficiency of Rollout-Free Training}
\begin{figure}[t]
    \centering
    \begin{subfigure}[t]{0.49\linewidth}
        \centering
        \includegraphics[width=\linewidth]{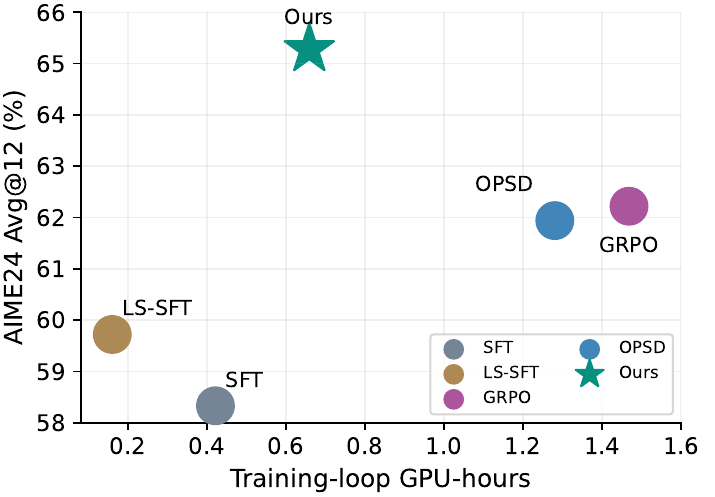}
        \caption{Accuracy--compute trade-off.}
        \label{fig:accuracy_compute}
    \end{subfigure}
    \hfill
    \begin{subfigure}[t]{0.49\linewidth}
        \centering
        \includegraphics[width=\linewidth]{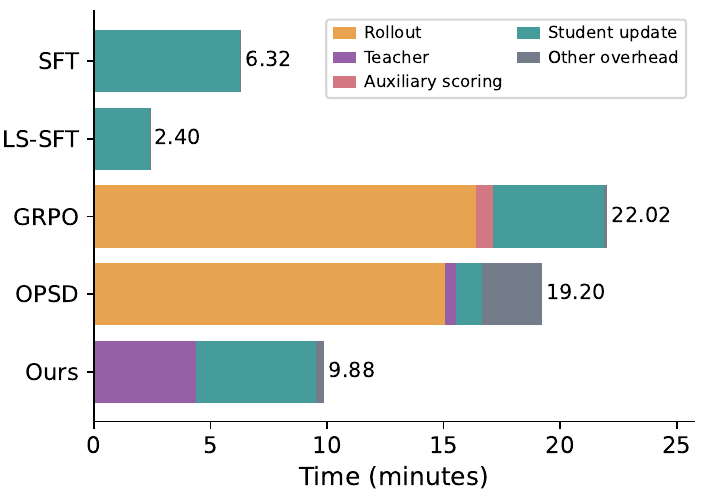}
        \caption{Runtime breakdown.}
        \label{fig:runtime_breakdown}
    \end{subfigure}

    \caption{
        \textbf{Training efficiency.}
        \textbf{(a)} AIME24 Avg@12 accuracy versus training-loop GPU-hours.
        \textbf{(b)} Runtime decomposed by computational component;
        numbers at the bar ends indicate total runtime.
    }
    \label{fig:efficiency}
\end{figure}

We now validate whether avoiding student rollouts provides the intended computational benefit.
Figure~\ref{fig:accuracy_compute} reports AIME24 avg@12 accuracy and training-loop cost across the evaluated methods.

\paragraph{Accuracy-compute trade-off.} 
Figure~\ref{fig:accuracy_compute} compares AIME24 Avg@12 against training-loop GPU-hours.
\model{} achieves the highest observed accuracy among the evaluated methods, reaching approximately $65.3\%$ with $0.66$ GPU-hours.
In comparison, OPSD and GRPO achieve approximately $61.9\%$ and $62.2\%$, while consuming $1.28$ and $1.47$ GPU-hours, respectively.
Thus, \model{} achieves slightly higher accuracy while reducing training-loop compute by approximately $49\%$ relative to OPSD and $55\%$ relative to GRPO.
Although SFT and LS-SFT consume less compute at their reported operating points, both remain below $60\%$ accuracy.
These results highlight the favorable accuracy--compute trade-off of \model{}: it achieves higher accuracy than the supervised baselines while requiring substantially less compute than the evaluated on-policy methods.

\paragraph{Runtime breakdown.}
To understand the source of these savings, Figure~\ref{fig:runtime_breakdown} decomposes the measured runtime into its constituent components.
Rollout generation dominates the runtime of both GRPO and OPSD.
In contrast, \model{} avoids student-generated rollouts, spending most of its runtime on teacher computation and student updates.
Notably, teacher computation is more expensive for \model{} than for OPSD in this breakdown, but this additional cost is more than offset by eliminating rollout generation.
The measured runtime of \model{} is $9.88$ minutes, compared with $19.20$ minutes for OPSD and $22.02$ minutes for GRPO, corresponding to speedups of $1.94\times$ and $2.23\times$, respectively.

Together with the accuracy comparison, these results demonstrate that rollout-free privileged distillation can achieve competitive reasoning performance while substantially reducing the computational cost of training.

\section{Related Work}
\subsection{On-Policy and Off-Policy Learning}
SFT learns from fixed reference trajectories~\citep{NEURIPS2022_b1efde53}, whereas on-policy methods train on model-generated contexts to address training--inference mismatch.
Examples include distillation methods such as GKD~\citep{agarwal2024onpolicy} and MiniLLM~\citep{ICLR2024_8ac015d4}, and reinforcement learning with verifiable rewards such as GRPO~\citep{shao2024deepseekmathpushinglimitsmathematical}.
More closely related, OPSD~\citep{zhao2026selfdistilled} and SDFT~\citep{shenfeld2026selfdistillation} use demonstration-conditioned teachers to supervise student-generated trajectories.
\model{} instead distills on fixed reference prefixes, improving their supervision without autoregressive rollout generation.

\subsection{Distributional Supervision}
Knowledge distillation conveys preferences beyond the observed label through teacher probabilities~\citep{hinton2015distillingknowledgeneuralnetwork}, while uniform label smoothing assigns equal mass to alternative tokens~\citep{3295222.3295349}.
Token-level distillation transfers next-token distributions; sequence-level distillation learns from generated outputs~\citep{kim-rush-2016-sequence}.
Self-distillation can also improve code generation by applying SFT to a model's own sampled solutions~\citep{zhang2026embarrassinglysimpleselfdistillationimproves}.
\model{} transfers context-dependent distributions at existing reference positions without generating replacement solutions.
Its teacher draws on information available only during training, following the privileged-information principle of generalized distillation~\citep{lopezpaz2016unifyingdistillationprivilegedinformation}.
Related work, $\pi$-Distill, jointly trains privileged and unprivileged policies using teacher-generated trajectories~\citep{penaloza2026privileged}; \model{} uses a frozen teacher conditioned on the current reference step.

\subsection{Local Alignment}
Process supervision associates feedback with intermediate reasoning decisions~\citep{uesato2022solvingmathwordproblems,ICLR2024_aca97732}, including automatically constructed step-level supervision in Math-Shepherd~\citep{wang-etal-2024-math}.
\model{} uses reference steps to construct next-token distributions without step-correctness annotations or a process reward model.
Step structure also informs on-policy distillation: SOD adjusts distillation strength using step-level divergence~\citep{zhong2026sodstepwiseonpolicydistillation}, StepOPSD combines hindsight-enriched teacher contexts with step-aware advantage shaping~\citep{zhang2026stepopsdstepawareonlinepreference}, and SOPD generates teacher continuations at step boundaries along student trajectories~\citep{sun2026steplevelonpolicydistillationinterpolating}.
In \model{}, step boundaries determine which privileged information constructs the target at each reference position, while preserving the student's complete reference history.
This aligns distributional supervision with the current reference step in a fully off-policy setting.
\section{Conclusion}
We introduced \model{}, a rollout-free self-distillation method that uses step-aligned privileged information to construct distributional targets on fixed reference prefixes. On mathematical reasoning benchmarks, \model{} outperforms SFT and label-smoothed SFT on average and remains competitive with GRPO and OPSD, while largely preserving out-of-domain coding performance and achieving approximately $2\times$ training-loop speedups over the on-policy baselines. Analyses support the value of context-dependent preferences and current-step guidance. These findings suggest that carefully constructed supervision can make fixed demonstrations a competitive and efficient basis for reasoning post-training.

\clearpage
% \subsection*{AI use statement}
% Generative AI tools were used to assist with language polishing and LaTeX formatting. The authors reviewed the resulting edits and take full responsibility for the accuracy and integrity of the final manuscript.

\bibliography{iclr2027_conference}
\bibliographystyle{iclr2027_conference}

\clearpage
\appendix
\section{Training Details}
\label{app:training_details}
We follow the baseline training settings in OPSD~\citep{zhao2026selfdistilled}, as summarized in Table~\ref{tab:method_configs}. LS-SFT follows the SFT configuration with label-smoothing coefficient $\epsilon_{\mathrm{ls}}=0.1$.

\begin{table}[t]
\centering
\caption{Training settings for baselines and \model{}. LS-SFT extends the SFT configuration with label smoothing. A dash denotes a setting that does not apply. Sequence limits refer to teacher-forced inputs; rollout lengths refer to generated responses.}
\label{tab:method_configs}
\small
\setlength{\tabcolsep}{4pt}
\renewcommand{\arraystretch}{1.15}
\begin{tabular}{lccccc}
\toprule
Setting & SFT & LS-SFT & GRPO & OPSD & \model{} \\
\midrule
Learning rate & $5\times10^{-6}$ & $5\times10^{-6}$ & $5\times10^{-6}$ & $5\times10^{-6}$ & $5\times10^{-6}$ \\
Effective batch size & 32 & 32 & 32 & 32 & 32 \\
LoRA rank & 64 & 64 & 64 & 64 & 64 \\
LoRA alpha & 128 & 128 & 128 & 128 & 128 \\
Training steps & 100 & 100 & 500 & 100 & 100 \\
Max sequence length & 16K & 16K & -- & -- & 16K \\
Rollout length & -- & -- & 16K & 1K & -- \\
Generations per prompt & -- & -- & 8 & 1 & -- \\
Sampling temperature & -- & -- & 1.2 & 1.1 & -- \\
Label-smoothing coefficient & 0 & 0.1 & -- & -- & -- \\
\bottomrule
\end{tabular}
\end{table}

\section{Evaluation Details}
\label{app:evaluation_details}
We follow the evaluation protocol of \citet{zhao2026selfdistilled}.
We evaluate all models with a sampling temperature of $1.0$ and top-$p$ of $0.8$, with top-$k$ filtering disabled (\texttt{top\_k=-1}).
We set the maximum generation length to 38k tokens and the maximum total sequence length, including the prompt and generated response, to 40k tokens.
For mathematical reasoning benchmarks, we follow the OPSD evaluation implementation\footnote{\url{https://github.com/siyan-zhao/OPSD}} for prompt construction, chat-template application, answer extraction, and correctness grading. Specifically, we apply each model's native chat template and instruct it to reason step by step and place its final answer within \texttt{\textbackslash boxed\{\}}.
We extract the last boxed answer and assess its correctness against the ground-truth answer using the implementation's \texttt{math\_verify}-based grading function.

\section{Teacher and Student Prompt Examples}
\label{app:prompt_templates}
We show a real example from the training split of \texttt{siyanzhao/Openthoughts\_math\_30k\_opsd}, asking for $y$ given $x^{2y}=4$ and $x=4$.
The blocks below show the user message and the open assistant prefix supplied to each branch. Role labels are explanatory; the model's native non-thinking chat template supplies the role delimiters. No additional system message is provided.

\begin{promptbox}{Student Prompt}
[User]
Problem: If $x^{2y}= 4$ and $x = 4$, what is the value of $y$? Express your answer as a common fraction.

Please reason step by step, and put your final answer within \boxed{}.

[Assistant prefix]
Let's solve the problem step by step.
Okay, so I need to find the value of y in the equation x^(2y) = 4, given that x is equal to 4. Let me start by writing down what I know.

\end{promptbox}
The student sees the question and the first reference segment, but not the second segment as privileged guidance. Its continuous reference sequence uses a blank line between segments.

\begin{promptbox}{Teacher Prompt}
[User]
Problem: If $x^{2y}= 4$ and $x = 4$, what is the value of $y$? Express your answer as a common fraction.

Potential next step:
First, the equation is x raised to the power of 2y, and that equals 4. Since they told me that x is 4, I can substitute that into the equation. Let me do that step by step.

Please reason step by step using the potential next step as guidance, and put your final answer within \boxed{}.

[Assistant prefix]
Let's solve the problem step by step.
Okay, so I need to find the value of y in the equation x^(2y) = 4, given that x is equal to 4. Let me start by writing down what I know.
\end{promptbox}
The teacher additionally sees the complete reference segment under ``Potential next step.'' Its assistant prefix contains the same prefix and ends with a single newline. Both branches score the same reference continuation beginning with ``First, the equation is x raised to the power of 2y''. At subsequent positions, the within-segment reference prefix is extended for both branches, while only the teacher retains access to the complete current segment as privileged guidance. Both branches use teacher forcing rather than sampling a continuation.

\end{document}